%% file: main_preprint.tex
\documentclass[conference]{IEEEtran}
\IEEEoverridecommandlockouts

\usepackage{cite}
\usepackage{amsmath,amssymb,amsfonts}
\usepackage{algorithmic}
\usepackage{graphicx}
\usepackage{textcomp}
\usepackage{tabularx}
\usepackage{vcell}
\usepackage{colortbl}
\usepackage{hhline}
\usepackage{ragged2e} 
\usepackage{booktabs}
\usepackage{placeins} 
\usepackage{multirow}   
\usepackage{siunitx}
\usepackage{makecell}
\usepackage{threeparttable}
\usepackage{array}
\usepackage{rotating}
\usepackage{cuted}
\usepackage{caption} 
\usepackage{pifont} 
\usepackage{float}
\usepackage{pifont} 

\newcommand{\cross}{\ding{55}}
\newcommand{\tick}{\ding{51}}
\newcommand{\eg}{{\it e.g. }}

\newcommand{\bv}[1]{\boldsymbol{#1}}

\newcommand{\bea}{\begin{eqnarray}}
\newcommand{\eea}{\end{eqnarray}}

\newcommand{\beas}{\begin{eqnarray*}}
\newcommand{\eeas}{\end{eqnarray*}}
\renewcommand{\baselinestretch}{0.98}

\def\BibTeX{{\rm B\kern-.05em{\sc i\kern-.025em b}\kern-.08em
    T\kern-.1667em\lower.7ex\hbox{E}\kern-.125emX}}
\begin{document}

\title{Learning Varying Physical Therapist-Patient Interactions for Robot-mediated Upper Limb Task-Specific Training}
\author{Jia Quan Loh\thanks{Jia Quan Loh, Vincent Crocher, Denny Oetomo and Ying Tan are with the Human Robotics Laboratory, Department of Mechanical Engineering, The University of Melbourne. Correspondence: \textcolor{blue}{jqloh@student.unimelb.edu.au}}, Vincent Crocher, Marlena Klaic, Denny Oetomo and Ying Tan
\thanks{Marlena Klaic is with the Melbourne School of Health Sciences, University of Melbourne.}
\thanks{This work was partially supported by the 2024 Graeme Clark Institute Cross-Faculty Seed Funding Scheme.}
\thanks{This work involved human subjects or animals in its research. Approval of all ethical and experimental procedures and protocols was granted by the Melbourne Health Human Research Ethics Committee under Approval No. 31992 on 25th June 2025.}
}
\maketitle

\begin{abstract}
Upper extremity motor function recovery is positively linked to Task-Specific Training (TST) and sufficient therapy dosage. Rehabilitation robots can increase TST dosage via controlled, repetitive treatment and free therapists to simultaneously manage other patients, but it has yet to demonstrate significant benefits over conventional treatment. This is potentially linked to inaccurate robotic representation of personalised physical therapist-patient interaction and lack of practice variability during TST. Hence, we advocate for robotic interventions that preserve the personalised physical therapist-patient interactions when delivering TST for patients across varying practise conditions. We propose a Learning-from-Demonstration framework using Task-Parameterised Gaussian Mixture Models (TPGMM) to learn personalised physical therapist-patient interaction in Task-Specific exercises, mapping patient joint kinematics to therapist-applied torques using few demonstrations. The model is generalised to reconstruct therapist torques in new task variations. The framework was evaluated on physical interactions from 14 mock ``therapist-patient” pairs over three tasks of increasing complexity, each with six variations. A benchmark comparison against a Look-Up Table was conducted. The results show both methods reproducing interactions in unseen task variations that deviate slightly from the actual interaction, with TPGMM slightly outperforming LUT. Both methods reproduced interactions that gets increasingly closer to the actual interaction as task complexity increases.
\end{abstract}


\input{Sections/1_introduction}
\input{Sections/2_problem_form}
\input{Sections/3_experimental_methods}
\input{Sections/4_results}

\input{Sections/5_discussion}

\bibliographystyle{IEEEtran}
\bibliography{Biblio/bibliography}

\end{document}

%% file: Sections/1_introduction.tex
\section{Introduction}
The improvement of upper extremity motor function is positively correlated with the inclusion of functionally orientated therapy or Task-Specific Training (TST) \cite{tot_efficacy_lee,tot_efficacy_lang} and sufficient therapy dosage \cite{dose_efficacy_schneider,dose_efficacy_lang}. Incorporating robotic interventions into TST enables a robot to deliver controlled, repetitive treatment to a patient, increasing the training dosage beyond what conventional one-to-one therapy allows, while freeing the therapist to manage multiple patients simultaneously \cite{phrhi_crocher}.

However, robotic delivery of TST faces challenges: robots must understand the task and environment a patient is practising in, and adapt their role as a ``therapist" when interacting with the patient \cite{hhi_johnson}. The challenge is compounded when robots are expected to deliver TST in highly dynamic environments analogous to real world settings \cite{vary_tot_chrungoo}, where numerous variables (e.g. different task positions, movement speeds, and objects) can be introduced in TST. 

These challenges prevent current robot-patient interaction models from accurately modelling what therapists really do during TST \cite{hhi_johnson}—specifically, the desired movements, forces, and communication of intent therapist use when practising a task with a patient. This may explain why reviews over 24 clinical studies \cite{tot_cho, robot_tot_rozevink} found robotic TST no substantially better than conventional TST, despite robots' ability to deliver larger training doses.

Of these challenges, we focus on the robotic delivery of the physical forces exchanged between therapist and patient in TST, forming the basis for robots to physically interact with patients. Rehabilitation robots are often programmed with predefined physical Human–Robot Interaction (pHRI) schemes (\eg support, resist, guide, assist-as-needed \cite{rbt_control_rehab_dalla,rbt_control_rehab_baster}), designed to replicate evidence-based therapist interventions that promote motor recovery after stroke. However, rehabilitation literature has increasingly advocated for personalised interventions catered towards the specific neurological characteristics and impairments of a patient \cite{personalise_micera}. Consequently, as patients demonstrate varying degrees of impairment, targeted interventions may fall outside the capabilities of the robot's in-built pHRI schemes. This creates a need for therapists to easily design new pHRI customised to each patients in-situ if patient needs are beyond what is pre-programmed in a robot \cite{lut_interaction_just}.


To address this need, an experimental rehabilitation approach proposed around 2016 \cite{gmm_interaction_maaref} but recently coined Robot-mediated physical Human-Human Interaction (RHHI) \cite{phrhi_vianello} was proposed. The therapist works through a robot to influence patient movement, and the robot directly captures the personalised interaction for future reproduction, removing the need to select predefined controllers to imitate the interaction. Pilot studies showed feasible reproduction of physical therapist-patient interaction \cite{lut_interaction_just}, and recently, improved patient recovery in lower limb joints over conventional therapy \cite{phrhi_kuccuktabak}.

\begin{table*}[!ht]
    \centering
    \caption{Learning Therapist Interaction Profiles from Demonstration--- Summary of literature.}
    \label{lfd_interaction_lit}
    \begin{tabularx}{1.\textwidth} { 
      >{\centering\arraybackslash}p{0.6cm} 
      >{\raggedright\arraybackslash}p{2.4cm} 
      >{\raggedright\arraybackslash}X
      >{\raggedright\arraybackslash}X
      >{\raggedright\arraybackslash}X
      >{\centering\arraybackslash}p{0.85cm}
        }
        \toprule
        Ref.  & RHHI Configuration    & Input & Learning Method  & Desired Output & Variation\\
        \midrule
        \cite{gmm_interaction_maaref} & T-R-P& Measured end-effector position & GMM  & End-effector force & \cross\\
        \midrule
        \cite{seds_interaction_martinez}  & T-R-R-P&  Measured end-effector force  & GMM-based non-linear system & End-effector force derivative & \cross\\
        \midrule
         \cite{gmm_interaction_fong} & T-R-P& Measured end-effector position & GMM-based Impedance model & End-effector force & \cross\\
        \midrule
        \cite{gmm_interaction_martinez} & T-R-R-P& Measured patient end-effector kinematics & GMM & Patient end-effector position & \cross\\
        \midrule
         \cite{pff_interaction_najafi} & T-R-P& Measured end-effector kinematics &  Potential Field Function & End-effector force  &  \cross\\
        \midrule
        \cite{lut_interaction_just}  & T-R-P& Measured joint positions & Look-Up Table  & Joint torques  & \cross\\ 
        \midrule
        \cite{gmm_interaction_sorkhabadi} & T-P  & Measured joint kinematics & GMM-based Impedance model & Joint torques &\cross\\
        \midrule
        \cite{lut_interaction_luciani} & T-R-P & Measured joint kinematics & Look-Up Table & Joint torques & \cross\\
        \midrule
        \cite{lauretti_dmp_movement_2017} & T-R-P & Time & DMP & End-effector positions & \tick\\
        \midrule
        \cite{lauretti_dmp_movement_2018} & T-R-P & Time & DMP & Joint positions & \tick\\
        \midrule
        \cite{loh_tpgmm_movement_2025} & T-P & Time & Task-Parameterised GMM & Joint positions & \tick\\
        \bottomrule
    \end{tabularx}
    \begin{tablenotes}
    \footnotesize
    \item T: Therapist, R: Robot, P: Patient, GMM: Gaussian Mixture Models, DMP: Dynamic Movement Primitives
    \end{tablenotes}
    \vspace{-2em}
\end{table*}

While current RHHI approaches allows for directly capturing and imitating personalised therapist actions, its efficacy for robot-assisted TST when variations are incorporated remain unknown. Clinical studies reviewed in \cite{robot_tot_rozevink} report a lack of practice variability (only four out of 17 studies report variability). Cognitive science theory and motor learning studies investigating the Variability of Practice hypothesis \cite{variability_learning_raviv,variability_learning_A_shea,variability_learning_B_shea} emphasise the importance of including varying task contexts (e.g. different reaching and grasping points, varying movement speeds) to improve motor skill learning. Lack of practice variability leads to movement-specific learning that does not induce the neuro-plastic changes necessary for skill transfer to similar tasks or varied settings \cite{tst_pollock}. Capturing the required physical interaction does not enable robots to understand its application across varying conditions; therefore, constant therapist presence is potentially required to redemonstrate the interaction.

Motivated by this limitation, we believe that to achieve practical RHHI for variable robot-assisted TST, the latent features of the physical interactions applied by the therapist along a patient's movement could be learned, from a few demonstrations, and the learned interaction generalised to variations of the task. In such scenario, a clinician could demonstrate a task for a given patient-task and further let the patient use the robot for structured mass repetition \cite{hhi_johnson} over varying task conditions.

Learning-from-Demonstration (LfD) provides a framework for modelling pHRI in varying environments using data from few demonstrations\cite{lfd_review_ravichandar}, which could be suitable for modelling such physical therapist-patient interaction. Literature on learning physical therapist-patient interactions using LfD, is summarised in Table \ref{lfd_interaction_lit}. From the table, the extent of methods that focus on learning physical forces in RHHI \cite{gmm_interaction_maaref,seds_interaction_martinez,gmm_interaction_fong,gmm_interaction_martinez,pff_interaction_najafi,lut_interaction_just,gmm_interaction_sorkhabadi,lut_interaction_luciani} when the task conditions are varied remains unexplored, with approaches only ``reproducing'' previously seen condition. Generalising demonstrations in upper limb rehabilitation to unseen variations were approached \cite{lauretti_dmp_movement_2017,lauretti_dmp_movement_2018,loh_tpgmm_movement_2025}, though restricted to learning only the movement of the task.

Lauretti \textit{et al.}, for example, used  Dynamic Movement Primitives (DMP) to learn task and joint space movement of healthy and impaired subjects in robot-assisted TST~\cite{lauretti_dmp_movement_2017,lauretti_dmp_movement_2018}. They achieved generalisation to different conditions with a 100\% task completion rate for different task positions and posture. However, the similarity of patient's movement features along with the representation of the intended therapist-patient forces by the robot were not quantified and evaluated.

Recently, we proposed to use DMP and Task-Parameterised GMMs (TPGMM) to learn joint movements across different task variations, and compared the ability of both models to preserve the movement features demonstrated by healthy individuals~\cite{loh_tpgmm_movement_2025}. It shows that TPGMM preserves the intended motion features better than DMPs in unseen task variations. However, this study lacks a robotic implementation and evaluation of the intended therapist-patient forces.

Building on this literature, in this work, we propose to learn the physical therapist-patient interaction and to generalise it to unseen task variations. We first collect physical therapist-patient interactions comprised of patient joint kinematics and therapist-applied joint torques, across variations of task-specific exercises. We propose a LfD framework, to learn and generalise these interactions across task variations, as a therapist–patient–task-specific interaction model. Finally, we evaluate this method against a Look Up Table (LUT) approach proposed by Luciani \textit{et al.}~\cite{lut_interaction_luciani} in an experiment with physiotherapy students.





%% file: Sections/2_problem_form.tex
\section{Framework for Learning Physical Patient--Therapist Interaction}
\label{lfd}

This section presents the proposed framework for learning physical therapist--patient interactions during task-specific rehabilitation. The therapist--patient interaction is first mathematically formulated using patient joint kinematics and therapist-applied joint torques. A Task-Parameterized Gaussian Mixture Model (TPGMM)-based learning framework is then developed to capture and generalize therapist interactions across varying task conditions. Finally, a conventional look-up table (LUT) method is described as the benchmark for evaluating the proposed approach.

We define the physical interaction between a therapist and a patient by the interaction vector $\boldsymbol{\gamma}_{t}$, given by
\begin{equation}
\boldsymbol{\gamma}_{t} = [\boldsymbol{q}_{t}, \boldsymbol{\dot{q}}_{t}, \boldsymbol{\tau}_{t}]~\in \mathbb{R}^{3\cdot D},~t\in [0,T_n],
\end{equation}
where $\boldsymbol{q}_{t}$, $\boldsymbol{\dot{q}}_{t}$, and $\boldsymbol{\tau}_{t}\in\mathbb{R}^{D}$ denote the patient joint positions, joint velocities, and therapist-applied joint torques, respectively, for a $D$-DOF musculoskeletal model at time $t$. A therapist--patient pair, indexed by $a\in[1,\ldots,A]$, performs task $e\in[1,\ldots,E]$ under each task variation $v\in[1,\ldots,V]$. Each variation is repeated $R$ times, indexed by $r\in[1,\ldots,R]$. Each trial is uniquely identified by the index $n=\{a,e,v,r\}$ and has a duration of $T_n$ seconds.

The interaction vector $\boldsymbol{\gamma}_{t}$ is sampled at a fixed sampling period $\Delta t$, producing the discrete interaction vector $\boldsymbol{\gamma}_{n,j}$, where $j$ denotes the sample index within trial $n$. Consequently, trial $n$ consists of $J_n=T_n/\Delta t$ interaction samples collected over its duration $T_n$. The sequence of interaction vectors from trial $n$ forms the interaction profile $\Gamma_n$, given by
\begin{equation}
 \Gamma_n = [\boldsymbol{q}_{n}, \boldsymbol{\dot{q}}_{n}, \boldsymbol{\tau}_{n}] \in \mathbb{R}^{J_n \times 3 \cdot D},
 \label{int_vec}
\end{equation}
where $\boldsymbol{q}_{n}$, $\boldsymbol{\dot{q}}_{n}$, and $\boldsymbol{\tau}_{n}\in \mathbb{R}^{J_n \times D}$ denote the patient joint position, joint velocity, and therapist-applied joint torque profiles, respectively, over the $J_n$ samples of trial $n$.

For a given therapist--patient pair $a$ performing task $e$, demonstrations are collected over all task variations and repetitions, resulting in $N=V\cdot R$ trials. The corresponding interaction profiles are stacked vertically to form the task demonstration set $\boldsymbol{\Gamma}$, containing a total of $\boldsymbol{J}=\displaystyle \sum_{n=1}^{N}J_n$ samples, given by
\begin{equation}
\boldsymbol{\Gamma} =  \begin{bmatrix} \Gamma_1 \\ \Gamma_2 \\ \vdots \\ \Gamma_N \end{bmatrix}  = [\boldsymbol{q}, \boldsymbol{\dot{q}}, \boldsymbol{\tau}] \in \mathbb{R}^{\boldsymbol{J} \times 3 \cdot D},
\end{equation}
where $\boldsymbol{q}$, $\boldsymbol{\dot{q}}$, and $\boldsymbol{\tau}\in \mathbb{R}^{\boldsymbol{J} \times D}$ denote the patient joint position, joint velocity, and therapist-applied joint torque profiles from all $N$ trials stacked vertically.

Following previous studies \cite{lut_interaction_luciani,lut_interaction_just,gmm_interaction_sorkhabadi}, we seek to learn, for each therapist--patient pair $a$ and task $e$, a therapist--patient--task-specific interaction model $g_{a,e}$ that maps the patient's joint kinematics $(\boldsymbol{q},\boldsymbol{\dot{q}})$ to the therapist-applied joint torques $\boldsymbol{\tau}$. A Learning-from-Demonstration (LfD) approach is employed to learn an approximation $\hat{g}_{a,e}$ of the unknown interaction model $g_{a,e}$. The estimated therapist-applied joint torques are approximated by
\begin{equation}
\hat{\boldsymbol{\tau}} = \hat{g}_{a,e}(\boldsymbol{q},\boldsymbol{\dot{q}}).
\end{equation}

\subsection{Task-Parameterised Gaussian Mixture Models}

In this work, we employ a Task-Parameterised Gaussian Mixture Model (TPGMM) to approximate the interaction model $\hat{g}_{a,e}$. TPGMM was selected because of its ability to generalise demonstrations across varying task conditions while preserving the key characteristics of the demonstrated interactions. In our previous work, TPGMM was shown to preserve the intended movement features more effectively than Dynamic Movement Primitives (DMPs) when reproducing unseen task variations, particularly as task complexity increases~\cite{loh_tpgmm_movement_2025}. TPGMM extends the standard Gaussian Mixture Model (GMM) by introducing a fixed number of task-dependent reference frames ($P$) for each demonstration. Each demonstration is first transformed with respect to these reference frames before a GMM is learned in each frame.



\subsubsection{Frame Transformation}

Each demonstration is assumed to consist of movements between a fixed number of $P$ task points (e.g., start, via, and end points). A reference frame is assigned to each task point, resulting in $P$ reference frames for every interaction profile $\Gamma_n$.

Following Calinon's formulation~\cite{tpgmm_calinon}, the $p^{th}$ reference frame of $\Gamma_n$ is defined by a rotation matrix $A_{n}^{(p)} \in \mathbb{R}^{D\times D}$ and a displacement vector $\boldsymbol{b}_{n}^{(p)} \in \mathbb{R}^{D}$ with respect to a known inertial frame. The interaction profile $\Gamma_n$ is subsequently transformed into each reference frame.

Specifically, the $j^{th}$ interaction vector $\boldsymbol{\gamma}_{n,j}$ is transformed into the $p^{th}$ reference frame to obtain $\boldsymbol{\gamma}^{(p)}_{n,j}$, given by
\begin{equation}
\begin{aligned}
\boldsymbol{\gamma}^{(p)}_{n,j} = (A_{n}^{(p)})^{-1} (\boldsymbol{\gamma}_{n,j}-\boldsymbol{b}_{n}^{(p)}),~\forall j\in [0,1,\dots,J_n].
\end{aligned}
\end{equation}

Applying this transformation to every interaction vector in trial $n$ yields the transformed interaction profile $\Gamma^{(p)}_n$. The transformed interaction profiles from all $N$ trials are then stacked to form the demonstration set in the $p^{th}$ reference frame,
\begin{equation}
\boldsymbol{\Gamma}^{(p)} = \begin{bmatrix} \Gamma_1^{(p)} \\ \Gamma_2^{(p)} \\ \vdots \\ \Gamma_N^{(p)} \end{bmatrix},~\forall p \in [1,2,\dots,P].
\end{equation}

\textbf{Note}: Unlike the conventional implementation of TPGMM in a three-dimensional Euclidean space, we apply TPGMM in the patient's $D$-DOF joint space \cite{loh_tpgmm_movement_2025}. Consequently, the reference frame parameters are given by
\begin{equation}
A_{n}^{(p)} = I_{D \times D}~and ~\boldsymbol{b}_{n}^{(p)} = [\boldsymbol{q}_{n,j_p},\boldsymbol{\dot{q}}_{n,j_p}, 0_{D}],
\end{equation}
where $\boldsymbol{q}_{n,j_p}$ and $\boldsymbol{\dot{q}}_{n,j_p}$ denote the joint positions and joint velocities, respectively, at sample $j_p$ of trial $n$, corresponding to the instant when the patient reaches the $p^{th}$ task point. Since the interaction is represented in the patient's joint space, no rotational or scaling transformation is tested, i.e., $A_{n}^{(p)} = I_{D \times D}$. Furthermore, the joint torque component is not translated between reference frames (represented by $0_D$ in $\boldsymbol{b}_{n}^{(p)}$), ensuring that demonstrations with identical joint kinematic configurations in different reference frames share the same joint torque profile. 

\subsubsection{Modelling}

A TPGMM models the demonstrations using a mixture of $K$ Gaussian kernels, denoted by $\{\pi_{k}, \mathcal{N}(\boldsymbol{\mu}_k,\Sigma_k)\}_{k=1}^{K}$. The $k^{th}$ Gaussian kernel is characterised by a mixture weight $\pi_{k}$ and a Gaussian distribution $\mathcal{N}(\boldsymbol{\mu}_k,\Sigma_k)$, where $\boldsymbol{\mu}_k\in\mathbb{R}^{D}$ and $\Sigma_k\in\mathbb{R}^{D\times D}$ denote the mean vector and covariance matrix, respectively.

To account for task variations, each Gaussian distribution is parameterised by the $P$ reference frames through a set of sub-Gaussian distributions, $\{\mathcal{N}(\boldsymbol{\mu}_k^{(p)},\Sigma_k^{(p)})\}_{p=1}^{P}$, where $\boldsymbol{\mu}_k^{(p)}$ and $\Sigma_k^{(p)}$ denote the mean vector and covariance matrix of the $k^{th}$ Gaussian kernel expressed in the $p^{th}$ reference frame. Consequently, the complete TPGMM is represented by
\[
\{\pi_{k}, \{\mathcal{N}(\boldsymbol{\mu}_k^{(p)},\Sigma_k^{(p)})\}_{p=1}^{P}\}_{k=1}^{K}.
\]

The optimal number of Gaussian kernels, $K$, is determined using the Bayesian Information Criterion (BIC)~\cite{bic_hastie}. For each reference frame $p$, a Gaussian Mixture Model (GMM), $\{\pi_{k}, \mathcal{N}(\boldsymbol{\mu}_k^{(p)},\Sigma_k^{(p)})\}_{k=1}^{K}$, is learned from the transformed demonstration set $\boldsymbol{\Gamma}^{(p)}$ using the Expectation--Maximisation (EM) algorithm. The EM algorithm estimates the mixture parameters by maximising the likelihood of the observed data $\boldsymbol{\Gamma}^{(p)}$ under the corresponding GMM~\cite{tpgmm_calinon}, for all $p\in[1,2,\dots,P]$.

\subsubsection{Regression}

To predict therapist-applied joint torques for a new task variation, a new set of reference frames is first defined by $\bv{b}_{new}^{(p)}$, corresponding to the required joint kinematics at the new task points. The learned GMM in each reference frame, $\{\pi_{k}, \mathcal{N}(\boldsymbol{\mu}_k^{(p)},\Sigma_k^{(p)})\}_{k=1}^{K}$, is then conditioned on the new reference frames to obtain
\[
\{\pi_{k}, \mathcal{N}(\hat{\boldsymbol{\mu}}_k^{(p)},\hat\Sigma_k^{(p)})\}_{k=1}^{K},
\]
given by
\begin{equation}
\boldsymbol{\hat \mu}_k^{(p)} = \boldsymbol{\mu}_k^{(p)} + \bv{b}_{new}^{(p)} ,~\hat\Sigma_k^{(p)} = \Sigma_k^{(p)}.
\label{eq:cond1}
\end{equation}

\textbf{Note}: Equation~\eqref{eq:cond1} is a simplified form of the conditioning proposed by Calinon~\cite{tpgmm_calinon}, since the transformation matrix in the joint space is the identity matrix, i.e., $A_{new}^{(p)}=I$.

The conditioned Gaussian distributions from all $P$ reference frames are subsequently combined using the product of Gaussians to obtain the reduced GMM,
\[
\Theta=\{\pi_{k}, \mathcal{N}(\hat{\boldsymbol{\mu}}_k,\hat\Sigma_k)\}_{k=1}^{K},
\]
where
\begin{equation}
\boldsymbol{\hat \mu}_k = \hat\Sigma_{k}\sum_{p=1}^P \hat\Sigma_{k}^{(p)^{-1}} \boldsymbol{\hat \mu}_k^{(p)},~\quad \hat \Sigma_k = \Biggl(\sum_{p=1}^P \hat\Sigma_{k}^{(p)^{-1}}\Biggl)^{-1}.
\label{eq:cond3}
\end{equation}

Finally, Gaussian Mixture Regression (GMR) is performed on the reduced GMM $\Theta$ to approximate the learned interaction model $\hat{g}_{a,e}$. Given a new patient joint kinematic input $(\boldsymbol{q}_{j_{\mathrm{test}}},\boldsymbol{\dot{q}}_{j_{\mathrm{test}}})$, the corresponding therapist-applied joint torques are estimated as
\begin{equation}
\boldsymbol{\hat\tau}_{j_{\mathrm{test}}} = \hat{g}_{a,e}(\boldsymbol{q}_{j_{\mathrm{test}}},\boldsymbol{\dot{q}}_{j_{\mathrm{test}}}).
\end{equation}

\subsection{Look-Up Table}

The Look-Up Table (LUT) method proposed by Luciani \textit{et al.}~\cite{lut_interaction_luciani} is implemented as the benchmark for evaluating the proposed TPGMM framework. Unlike TPGMM, which learns a statistical model to generalise therapist--patient interactions across varying task conditions, the LUT directly stores demonstrated interactions and predicts therapist-applied joint torques by retrieving the demonstrated interaction with the most similar joint kinematics.

Given the task demonstration set $\boldsymbol{\Gamma}=[\boldsymbol{q},\boldsymbol{\dot{q}},\boldsymbol{\tau}]$, comprising all patient joint position, joint velocity, and therapist-applied joint torque samples from therapist--patient pair $a$ performing task $e$, a LUT is constructed.

Given a new patient joint kinematic input $(\boldsymbol{q}_{j_{\mathrm{test}}},\boldsymbol{\dot{q}}_{j_{\mathrm{test}}})$, a closest-vector search is performed over $\boldsymbol{\Gamma}$ to identify the demonstration sample with the smallest Euclidean distance to the input. The therapist-applied joint torque associated with the nearest neighbour is then selected as the estimated therapist-applied joint torque, $\boldsymbol{\hat{\tau}}_{j_{\mathrm{test}}}$.

To reduce rapid variations in the estimated torque caused by switching between neighbouring LUT entries, $\boldsymbol{\hat{\tau}}_{j_{\mathrm{test}}}$ is subsequently filtered using a low-pass filter with a cut-off frequency of $3$ Hz~\cite{lut_interaction_luciani}.

%% file: Sections/3_experimental_methods.tex
\section{Experimental Methods}

This section describes the experimental methodology used to evaluate the proposed therapist--patient interaction learning framework. The experimental protocol, participant recruitment, data acquisition system, and data processing pipeline are first presented. The training, validation, and generalisation procedures are then described, followed by the outcome measures and statistical analyses used to compare the proposed TPGMM framework with the benchmark LUT approach.

\subsection{Experimental Protocol}
\subsubsection{Experimental Procedure}
Pairs of participants were recruited for the study to role play as both mock patient and therapist. In each half-session, one participant is instructed to act out three ``patient'' personas each performing one Task-specific exercise. The other participant interacts with the ``patient'' based on their clinical expertise as the ``therapist''. At the end of each half-session, the two participants exchange roles.  

\begin{figure}[!htbp]
    \centering
    \includegraphics[width=1\linewidth]{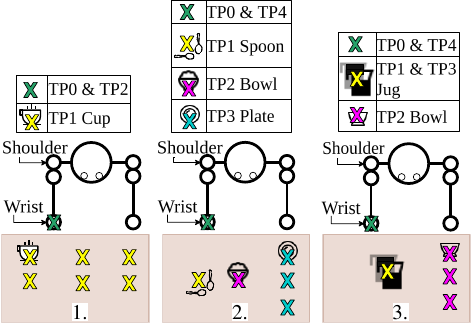}
    \caption{Task-specific exercises performed in the study. TP refers to task points where the joint kinematics are extracted from for the task frames. \textcolor{blue}{1.}, \textcolor{blue}{2.} and \textcolor{blue}{3.} refers to each patient persona.}
    \vspace{-1em}
    \label{tasks}
\end{figure}

Each task consists of a series of sub-tasks performed under specific variations. The persona diagnosis and sub-tasks are defined in Table \ref{sbmvmt}. Each task has two parameters for variations, one with two levels, one with three levels, as described in Table \ref{sbmvmt}. Each variation of each task is repeated for four trials. An illustration of the tasks is given in Fig \ref{tasks}.

\begin{table*} [htbp!]
\centering
\setlength{\extrarowheight}{1pt}  
\setlength{\tabcolsep}{2pt}
\caption{Patient personas (P$\#$) with associated task descriptions.}
\arrayrulecolor{black}
\arrayrulecolor{black}
\begin{tabular}{
>{\raggedright\arraybackslash}p{0.4cm}
>{\raggedright\arraybackslash}p{1.4cm}
>{\raggedright\arraybackslash}p{3.9cm}
l
>{\raggedright\arraybackslash}p{1.6cm}
>{\raggedright\arraybackslash}p{2.1cm}
>{\raggedright\arraybackslash}p{2cm}
l
>{\raggedright\arraybackslash}p{0.4cm}
>{\raggedright\arraybackslash}p{5cm}
} \cline{2-10}
 & \multicolumn{2}{c}{Patient History} &  & \multicolumn{3}{c}{Task Description} &  & \multicolumn{2}{c}{Sub-tasks} \\[2pt] \hhline{~--~---~--}
P1 & {\cellcolor[rgb]{0.753,0.753,0.753}}FMA-UE & 8/66 &  & {\cellcolor[rgb]{0.753,0.753,0.753}}Task & \multicolumn{2}{l}{Reaching towards a cup and return} &  & {\cellcolor[rgb]{0.753,0.753,0.753}}1-1 & Move wrist from home (TP0) to cup (TP1) \\ \hhline{~--~---~>{\arrayrulecolor[rgb]{0.753,0.753,0.753}}-~}
 & {\cellcolor[rgb]{0.753,0.753,0.753}}MMSE & 28/30 &  & {\cellcolor[rgb]{0.753,0.753,0.753}}Variations & Cup Distance & Close &  & {\cellcolor[rgb]{0.753,0.753,0.753}}1-2 & Move wrist from cup (TP1) to home (TP2) \\ \hhline{~>{\arrayrulecolor{black}}--~>{\arrayrulecolor[rgb]{0.753,0.753,0.753}}-~~~>{\arrayrulecolor{black}}--}
 & {\cellcolor[rgb]{0.753,0.753,0.753}}Diagnosis & Dense right hemiparesis &  & {\cellcolor[rgb]{0.753,0.753,0.753}} &  & Far &  &  &  \\ \hhline{~>{\arrayrulecolor[rgb]{0.753,0.753,0.753}}-~~->{\arrayrulecolor{black}}--~~~}
 & {\cellcolor[rgb]{0.753,0.753,0.753}} & Left middle cerebral artery stroke (6 weeks) &  & {\cellcolor[rgb]{0.753,0.753,0.753}} & Shoulder Rotation & External rotation &  &  &  \\ \hhline{~--~>{\arrayrulecolor[rgb]{0.753,0.753,0.753}}-~~~~~}
 & {\cellcolor[rgb]{0.753,0.753,0.753}}Presentation & Severe weakness in right upper extremity &  & {\cellcolor[rgb]{0.753,0.753,0.753}} &  & No rotation &  &  &  \\
 & {\cellcolor[rgb]{0.753,0.753,0.753}} & Cannot initiate active grasp and voluntary movement against gravity &  & {\cellcolor[rgb]{0.753,0.753,0.753}} &  & Internal rotation &  &  &  \\ \arrayrulecolor{black}\cline{2-3}\cline{5-7}
\\[-0.1cm]
\hhline{~--~---~--}
P2 & {\cellcolor[rgb]{0.753,0.753,0.753}}FMA-UE & 32/66 &  & {\cellcolor[rgb]{0.753,0.753,0.753}}Task & \multicolumn{2}{l}{Using utensils to transport food} &  & {\cellcolor[rgb]{0.753,0.753,0.753}}2-1 & Move wrist from home (TP0) to hold spoon (TP1) \\ \hhline{~--~---~>{\arrayrulecolor[rgb]{0.753,0.753,0.753}}-~}
 & {\cellcolor[rgb]{0.753,0.753,0.753}}MMSE & 30/30 &  & {\cellcolor[rgb]{0.753,0.753,0.753}}Variations & Spoon Orientation & Upright &  & {\cellcolor[rgb]{0.753,0.753,0.753}}2-2 & Move spoon (TP1) to rice bowl (TP2) and scoop \\ \hhline{~>{\arrayrulecolor{black}}--~>{\arrayrulecolor[rgb]{0.753,0.753,0.753}}-~~~-~}
 & {\cellcolor[rgb]{0.753,0.753,0.753}}Diagnosis & Post-ischaemic stroke (3 months) &  & {\cellcolor[rgb]{0.753,0.753,0.753}} &  & Sideways &  & {\cellcolor[rgb]{0.753,0.753,0.753}}2-3 & Move spoon (TP2) to plate (TP3) and finish pouring \\ \hhline{~-~~->{\arrayrulecolor{black}}--~>{\arrayrulecolor[rgb]{0.753,0.753,0.753}}-~}
 & {\cellcolor[rgb]{0.753,0.753,0.753}} & Moderate motor return &  & {\cellcolor[rgb]{0.753,0.753,0.753}} & Plate Distance & Close &  & {\cellcolor[rgb]{0.753,0.753,0.753}}2-4 & Move spoon (TP3) to home (TP4) \\ \hhline{~>{\arrayrulecolor{black}}--~>{\arrayrulecolor[rgb]{0.753,0.753,0.753}}-~~~>{\arrayrulecolor{black}}--}
 & {\cellcolor[rgb]{0.753,0.753,0.753}}Presentation & Right-sided weakness &  & {\cellcolor[rgb]{0.753,0.753,0.753}} &  & Middle &  &  &  \\
 & {\cellcolor[rgb]{0.753,0.753,0.753}} & Can initiate shoulder and elbow movement &  & {\cellcolor[rgb]{0.753,0.753,0.753}} &  & Far &  &  &  \\ \hhline{~>{\arrayrulecolor[rgb]{0.753,0.753,0.753}}-~~>{\arrayrulecolor{black}}---~~~}
 & {\cellcolor[rgb]{0.753,0.753,0.753}} & Limited wrist extension and fine motor control &  &  &  &  &  &  &  \\ \cline{2-3}
\\[-0.1cm]
\hhline{~--~---~--}
P3 & {\cellcolor[rgb]{0.753,0.753,0.753}}FMA-UE & 56/66 &  & {\cellcolor[rgb]{0.753,0.753,0.753}}Task & \multicolumn{2}{l}{Pour water from a jug} &  & {\cellcolor[rgb]{0.753,0.753,0.753}}2-1 & Move wrist from home (TP0) to hold jug (TP1) \\ \hhline{~--~---~>{\arrayrulecolor[rgb]{0.753,0.753,0.753}}-~}
 & {\cellcolor[rgb]{0.753,0.753,0.753}}MMSE & 30/30 &  & {\cellcolor[rgb]{0.753,0.753,0.753}}Variations & Jug Volume & Half Full &  & {\cellcolor[rgb]{0.753,0.753,0.753}}2-2 & Move jug (TP1) to rice bowl (TP2) and finish pouring set amount \\ \hhline{~>{\arrayrulecolor{black}}--~>{\arrayrulecolor[rgb]{0.753,0.753,0.753}}-~~~-~}
 & {\cellcolor[rgb]{0.753,0.753,0.753}}Diagnosis & Mild right hemiparesis post-lacunar infarct &  & {\cellcolor[rgb]{0.753,0.753,0.753}} &  & Full &  & {\cellcolor[rgb]{0.753,0.753,0.753}}2-3 & Move jug (TP2) and release at original jug position (TP3) \\ \hhline{~-~~->{\arrayrulecolor{black}}--~>{\arrayrulecolor[rgb]{0.753,0.753,0.753}}-~}
 & {\cellcolor[rgb]{0.753,0.753,0.753}} & Shoulder hitching and synergistic movements &  & {\cellcolor[rgb]{0.753,0.753,0.753}} & Bowl Distance & Close &  & {\cellcolor[rgb]{0.753,0.753,0.753}}2-4 & Move wrist (TP3) to home (TP4) \\ \hhline{~>{\arrayrulecolor{black}}--~>{\arrayrulecolor[rgb]{0.753,0.753,0.753}}-~~~>{\arrayrulecolor{black}}--}
 & {\cellcolor[rgb]{0.753,0.753,0.753}}Presentation & Mild coordination deficits and slight delay in fine control &  & {\cellcolor[rgb]{0.753,0.753,0.753}} &  & Middle &  &  &  \\
 & {\cellcolor[rgb]{0.753,0.753,0.753}} & Elevated shoulder compensation noted &  & {\cellcolor[rgb]{0.753,0.753,0.753}} &  & Far &  &  &  \\ \hhline{~--~---~~~}
\end{tabular}
\label{sbmvmt}
\vspace{-1.5em}
\end{table*}

A therapist--patient pair $a \in \{1,\dots,A\}$, demonstrates three task $e \in \{1,2,3\}$ for six variations $v \in \{1,2,\dots,6\}$ with four repetitions, $r \in \{1,2,\dots,4\}$ each.

\subsubsection{Participant Recruitment}
Participants who are third-year physiotherapy students that had underwent clinical placements in neuro-rehabilitation or sports therapy, were recruited from the Department of Physiotherapy, University of Melbourne. All procedures were approved by the Melbourne Health Human Research Ethics Committee ($\#31992$), and written informed consent was obtained prior the investigations.

\subsection{Data Collection}
\begin{figure*} [htbp!]
    \centering
    \includegraphics[width=0.95\linewidth]{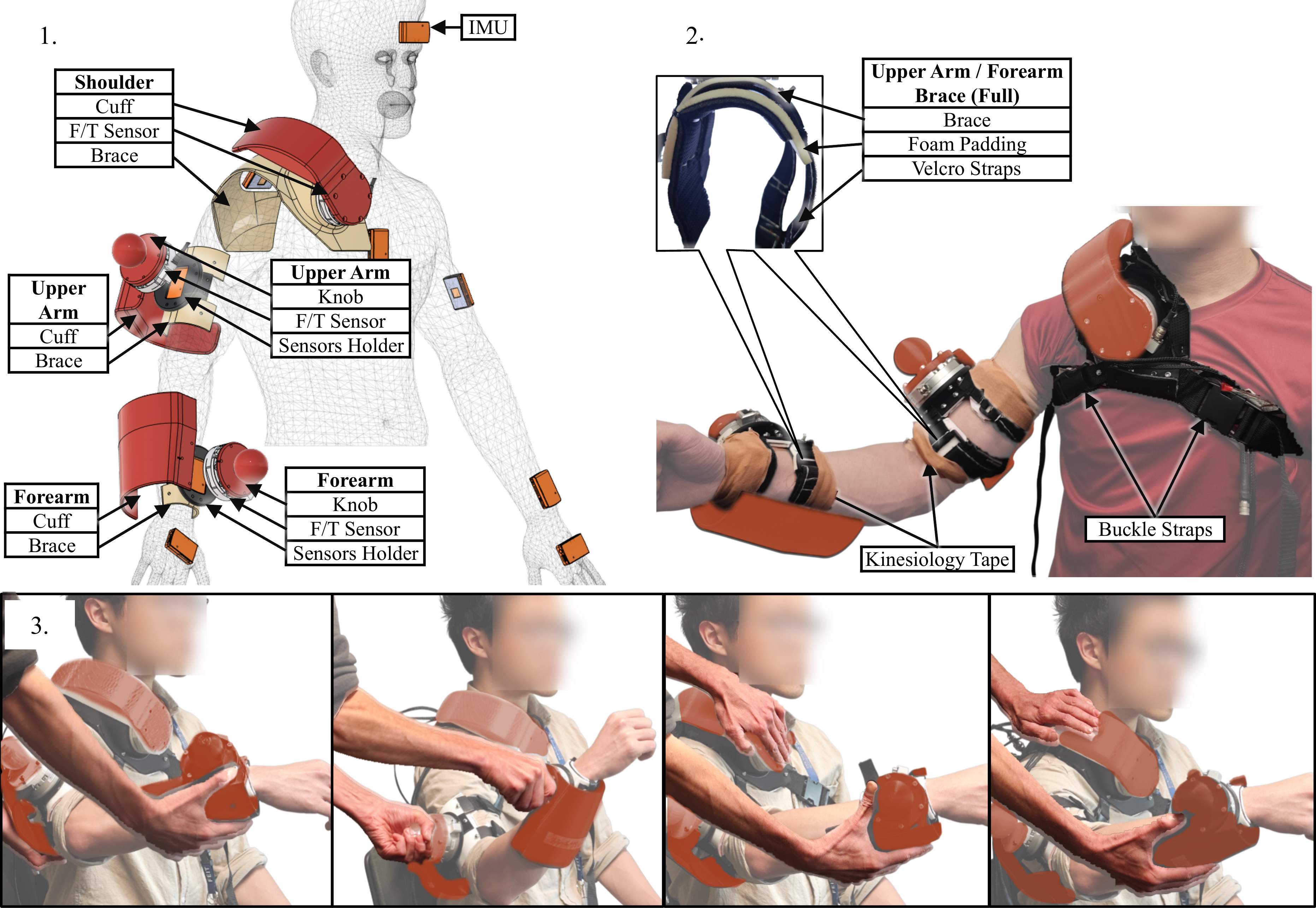}
    \caption{\textcolor{blue}{1.} Construction of TPU braces - PETG cuffs sensorised to measure force-torque interactions and placement of sensors on the wearer's body. Sensors holders house the IMU trackers on the right upper and forearm, with the force-torque sensors fixed on top. \textcolor{blue}{2.} Tight attachments on the wearer's body using buckle-velcro straps and kinesiology tapes. Foam padding is layered underneath the brace to increase surface contact with the skin. \textcolor{blue}{3.} Common interactions ``therapist" apply on the ``patient" when constrained to only interact with the sensorised cuffs (highlighted in red), identified during the experiment.}
    \label{ubo}
    \vspace{-1em}
\end{figure*}
In order to obtain force interaction measures during therapist-patient interaction, a wearable sensor system to record the wrenches exerted by the therapist on the patient's upper body was first designed. The overall system hardware design is illustrated in Fig.~\ref{ubo}.


The upper-body joint positions of the patient, $\boldsymbol{q}_{n,j} \in \mathbb{R}^{D=10}$, is measured using the XSENS Awinda motion capture system (XSENS, Netherlands). The upper body (pelvis to wrist) is here modelled as a four-link, 10 DOFs, serial manipulator after XSENS's MVN Biomechanical Model~\cite{XSENS}. 

The interaction wrenches between the patient upper-body and the therapist hands are measured by three 6-DOF RFT80-6A01 force-torque sensors (Robotous, Seongnam, South Korea) attached to custom rigid cuffs and placed on the shoulder, upper-arm and forearm (see Fig.~\ref{ubo}). For a given sample $j$, of trial $n$, each measured wrench is denoted $\boldsymbol{w}_{i,n,j} \in \mathbb{R}^{6}$ and $\{\boldsymbol{w}_{i,n,j}\}_{i=1,2,3}$ represents the wrenches sample of all three force sensors on the upper-body.


$\boldsymbol{q}_{n,j}$ and $\{\boldsymbol{w}_{i,n,j}\}_{i=1,2,3}$ are streamed into a Python application to synchronise both data on a common time frame and then resampled at $100Hz, \Delta t=0.01s$. Each trial $n$ provides joint positions and wrenches profile $\boldsymbol{q}_{n}$ and $\{\boldsymbol{w}_{i,n}\}_{i=1,2,3}$.

\subsection{Data Processing}

\subsubsection{Approximating Patient Joint Kinematics and Therapist-contributed Joint Torque }
$\boldsymbol{q}_{n}$ and $\{\boldsymbol{w}_{i,n}\}_{i=1,2,3}$  are filtered with a fourth-order low-pass filter cut-off at $10 Hz$. 

Two forms of biases were identified and subtracted from $\{\boldsymbol{w}_{i,j}\}_{i=1,2,3}$.  Wrench bias $\{\boldsymbol{w}_{i,\mathrm{bias}}\}_{i=1,2,3}$ is assumed as a result of the gradual deformation of the wearer's soft tissue. $\{\boldsymbol{w}_{i,\mathrm{bias}}\}_{i=1,2,3}$ is collected with the wearer held stationary and averaged over five seconds before the start of each variation. Gravity bias $\{\boldsymbol{w}_{i,\mathrm{gravity}}\}_{i=1,2,3}$ resulting from the mass of the 3D-printed cuffs are computed using the orientation of the IMU trackers and the weight of the cuffs.  

From Eq. \ref{int_vec}, an interaction profile was defined as $\Gamma_{n} = [\boldsymbol{q}_{n}, \boldsymbol{\dot{q}}_{n}, \boldsymbol{\tau}_{n}]$. $\boldsymbol{\dot{q}}_{n}$ is approximated numerically using the forward difference of $\boldsymbol{q}_{n}$. The resulting derivative is filtered further using a fourth-order low-pass filter cut-off at $1.5 Hz$ to remove the residual high frequency component amplified when taking the derivative. 

$\boldsymbol{\tau}_{i,n}$ denoting the resultant joint torque profile from each $i^{th}$ interaction point, is computed using the $i^{th}$ force sensor's Jacobian $H_{i,n}$, and summed to obtain $\boldsymbol{\tau}_{n}$, given as
\begin{equation}
\boldsymbol{\tau}_{i,n}=(H_{i,n})^{T}\boldsymbol{w}_{i,n},~
\boldsymbol{\tau}_{n}=\sum_{i=1}^{3} \boldsymbol{\tau}_{i,n}~.
\end{equation}

\subsubsection{Segmentation and Alignment}
Each $\Gamma_{n}$ is segmented by the sub-task phases described in Table \ref{sbmvmt} via the sub-movement onset at the specified task points (TP in Fig.~\ref{tasks}), referencing the exact time stamps of the video recording in the experiment. Joint kinematics at TPs are extracted to provide the reference frames $\boldsymbol{b}_{n}^{(p)}$ for Section \ref{lfd}'s learning framework. 

To account for variability in interaction speed and duration across repetitions of a variation, we performed temporal alignment between trials of a variation. For all trials in each therapist-patient-task-variation $\boldsymbol{\Gamma}_{\zeta = \{a,e,v\}} = \Gamma_{n=\{\zeta,r\}},~\forall r \in [1,\dots,4]$, the joint kinematics in each trial are aligned to the same number of samples $J_\zeta$ using Dynamic Time Warping (DTW) \cite{dtw_sempena}, and the subsequent warping paths used to extract the corresponding joint torques. A random trial in $\boldsymbol{\Gamma}_{\zeta}$ is chosen as the reference signal. All variations of a therapist-patient-task $\boldsymbol{\Gamma} = \boldsymbol{\Gamma}_{\zeta=\{a,e,v\}},~\forall v \in [1,\dots,6]$ after alignment are normalised to a common $[0,1]$ time scale. 

\subsection{Generalisation (Variation Split) and Validation (Monte Carlo Sampling)}
From the six task variations per persona, we selected two variations as unseen variations to evaluate the generalisation capability of the algorithms. The therapist-patient-task-specific interaction model $g_{a,e}$ was trained on four variations, and generalised to the two unseen variations. To ensure sufficient coverage of the interaction space for training, only sets of two unseen variations differing by at least one level in both parameters were considered. This constraint yielded six valid train-generalisation combinations per $\{a,e\}$.

We validated the performance of LfD when reconstructing interaction for seen variations by randomly holding out one trial from each of the four training variations and evaluate $g_{a,e}$ against these held-out trials. We perform four Monte Carlo samplings for each combination to ensure each trial in the training set appears in the validation set once.

We trained a separate $g_{a,e}$ model across all combinations-samplings and evaluated it against the held-out trials and unseen variations, ensuring complete coverage of train-validation-generalisation combinations.

\subsection{Outcome Measures}
The variability of therapist behaviour in a variation $\boldsymbol{\tau}_{\zeta= \{a,e,v\}} = \boldsymbol{\tau}_{n=\{\zeta,r\}},~\forall r \in [1,\dots,4]$ is quantified by the central behaviour (\eg mean) and the behaviour boundaries of the therapist (\eg 95\% confidence interval) \cite{lut_interaction_just}. Using joint kinematic profiles $\boldsymbol{q}_{n},\boldsymbol{\dot{q}}_{n}$ from trials that were held out or in the unseen variations, the reconstructed torque profile, $\boldsymbol{\hat\tau}_{n} = \hat g_{a,e}(\boldsymbol{q}_{n},\boldsymbol{\dot{q}}_{n})$ is consistent to the therapist's intended interaction in that variation if $\boldsymbol{\hat\tau}_{n}$ lies within $\boldsymbol{\tau}_{\zeta}$. 

However, given the limited number of repetitions of a variation, we opt for a non-parametric version of the interpretation used by Just~\cite{lut_interaction_just}. The central behaviour in each $\boldsymbol{\tau}_{\zeta}$ is instead  represented by the therapist's mid-range sample $\boldsymbol{m}_{\zeta,j}$ at each $j^{th}$ sample and peak torque $\boldsymbol{m}_{\zeta,\mathrm{peak}}$, given as
\begin{equation}
\begin{aligned}
\label{mid-range}
\boldsymbol{m}_{\zeta,j} = \frac{\max\limits_{r=1,2,3,4}{\{\boldsymbol{\tau}_{n,j}\}+\min\limits_{r=1,2,3,4}{\{\boldsymbol{\tau}_{n,j}\}}}}{2},\\
\forall j\in[0,\dots,J_\zeta],
\end{aligned}
\end{equation}
\begin{equation}
\boldsymbol{m}_{\zeta,\mathrm{peak}} = \frac{\max\limits_{r=1,2,3,4}{\{\boldsymbol{\tau}_{n,\mathrm{peak}}\}}+\min\limits_{r=1,2,3,4}\{{\boldsymbol{\tau}_{n,\mathrm{peak}}\}}} {2}
~.
\end{equation}

The reconstructed torques $\boldsymbol{\hat\tau}_n$, are evaluated against the variability of therapist behaviour using these outcome measures:
\begin{enumerate}
    \item \textit{Normalised average reconstruction error $\boldsymbol{\epsilon}_{n}$}
        \begin{equation}
        \boldsymbol{\epsilon}_{n} = \frac{1}{J_\zeta} \sum^{J_\zeta}_{j=0} \frac{|\boldsymbol{\hat\tau}_{n,j}-\boldsymbol{m}_{\zeta,j}|} {\max\limits_{r=1,2,3,4}{\{\boldsymbol{\tau}_{n,j}\}}-\boldsymbol{m}_{\zeta,j}};
        \end{equation}
    \item \textit{Normalised peak torque difference $\Delta\boldsymbol{\tau}_{n,\mathrm{peak}}$}\\
        \begin{equation}
        \Delta\boldsymbol{\tau}_{n,\mathrm{peak}} = \frac{|\boldsymbol{\hat\tau}_{n,\mathrm{peak}}-\boldsymbol{m}_{\zeta,\mathrm{peak}}|} {\max\limits_{r=1,2,3,4}{\{\boldsymbol{\tau}_{n,\mathrm{peak}}\}}-\boldsymbol{m}_{\zeta,\mathrm{peak}}}~and
        \end{equation}
    \item \textit{Coverage} $C_n~(\%)$ 
        \bea
        \boldsymbol{c}_{n,j} &=&\begin{cases}
        1 & \text{if } \boldsymbol{\hat\tau}_{n,j} \in [\min\limits_{r=1,2,3,4}, \max\limits_{r=1,2,3,4}]\{\boldsymbol{\tau}_{n,j}\},\\
        0 & \text{otherwise}.
        \end{cases} \hspace{0.2in}
        \\
            C_n &=& \frac{1}{J_\zeta} \sum^{J_\zeta}_{j=0} \boldsymbol{c}_{n,j}~.
        \eea
\end{enumerate}

We interpret $\boldsymbol{\epsilon}_{n}$ and $\Delta\boldsymbol{\tau}_{n,\mathrm{peak}}$ as how far away on average ${\boldsymbol{\hat\tau}}_n$ are from the central behaviour of the therapist in each variation, normalised by the behaviour boundaries. Values below $1$ indicate that ${\boldsymbol{\hat\tau}}_n$ stays within the boundaries of $\boldsymbol{\tau}_{\zeta}$, producing physical inputs consistent with the intended therapist interaction under the seen / unseen variation. We interpret $C_n$ as how long on average $\boldsymbol{\hat\tau}_n$ stays within $\boldsymbol{\tau}_{\zeta}$ as a percentage of the normalised time.

\subsection{Statistical Analysis}
Outcome measures values were averaged across the ten different DOFs and subsequently all samples of all train-validation-generalisation combinations, resulting in one sample per candidate method per patient persona per outcome measure for each participant in both the validation (seen variations) and generalisation (unseen variations) cases. Outliers, outside of 1.5IQR, were removed. 


After testing distributions for normality, each outcome measure was analysed with a two-way repeated-measures ANOVA testing the effect of the patient personas, candidate reconstruction methods and interaction. The Greenhouse-Geisser correction was applied where appropriate based on Mauchly’s test of sphericity. For each outcome measure, the validation and generalisation cases were analysed separately, leading to six separate ANOVAs. Where appropriate, post-hoc analysis were performed to test the independent effects of personas and candidate methods (Holm correction applied to reduce false negatives risk). 



%% file: Sections/4_results.tex
\section{Results}
This section presents the experimental results of the proposed therapist--patient interaction learning framework. The reconstruction performance of the proposed TPGMM framework is evaluated and compared with the benchmark LUT approach for both seen and unseen task variations using the defined outcome measures and statistical analyses.
\begin{table*} [htbp!]
\centering
\setlength{\extrarowheight}{1pt}  
\setlength{\tabcolsep}{2pt} 
\caption{\textcolor{blue}{Top:} Repeated-measures ANOVA for Effects of Persona, Method and Interaction. \textcolor{blue}{Bottom:} Holm-corrected pairwise comparisons between independent within-subject factor levels. (Val: Seen Variations, Gen: Unseen Variations) }
\vspace{-0.5em}
\arrayrulecolor{black}
\begin{tabular}{cccrccrrllrccrrllrccrrr} \cline{4-23}
 &  &  & \multicolumn{20}{c}{RM-ANOVA} \\ \cline{4-23}
 &  &  & \multicolumn{6}{c}{Persona} & \multicolumn{1}{c}{} & \multicolumn{6}{c}{Method} & \multicolumn{1}{c}{} & \multicolumn{6}{c}{Persona * Method} \\ \cline{1-2}\cline{4-9}\cline{11-16}\cline{18-23}
Measure & Case &  & \multicolumn{1}{c}{SS} & df1 & df2 & \multicolumn{1}{c}{MS} & \multicolumn{1}{c}{F} & \multicolumn{1}{c}{P-val} & \multicolumn{1}{c}{} & \multicolumn{1}{c}{SS} & df1 & df2 & \multicolumn{1}{c}{MS} & \multicolumn{1}{c}{F} & \multicolumn{1}{c}{P-val} & \multicolumn{1}{c}{} & \multicolumn{1}{c}{SS} & df1 & df2 & \multicolumn{1}{c}{MS} & \multicolumn{1}{c}{F} & \multicolumn{1}{c}{P-val} \\ \cline{1-2}\cline{4-9}\cline{11-16}\cline{18-23}
\multirow{2}{*}{$\varepsilon$} & Val &  & 0.348 & 2 & 26 & 0.174 & 15.59 & 0.0*** &  & 0.133 & 1 & 13 & 0.133 & 7.623 & 0.016* &  & 0.003 & 2 & 26 & 0.002 & 0.162 & 0.766 \\
 & {\cellcolor[rgb]{0.753,0.753,0.753}}Gen &  & {\cellcolor[rgb]{0.753,0.753,0.753}}0.612 & {\cellcolor[rgb]{0.753,0.753,0.753}}2 & {\cellcolor[rgb]{0.753,0.753,0.753}}26 & {\cellcolor[rgb]{0.753,0.753,0.753}}0.306 & {\cellcolor[rgb]{0.753,0.753,0.753}}7.965 & {\cellcolor[rgb]{0.753,0.753,0.753}}0.002** &  & {\cellcolor[rgb]{0.753,0.753,0.753}}0.164 & {\cellcolor[rgb]{0.753,0.753,0.753}}1 & {\cellcolor[rgb]{0.753,0.753,0.753}}13 & {\cellcolor[rgb]{0.753,0.753,0.753}}0.164 & {\cellcolor[rgb]{0.753,0.753,0.753}}8.134 & {\cellcolor[rgb]{0.753,0.753,0.753}}0.014* &  & {\cellcolor[rgb]{0.753,0.753,0.753}}0.013 & {\cellcolor[rgb]{0.753,0.753,0.753}}2 & {\cellcolor[rgb]{0.753,0.753,0.753}}26 & {\cellcolor[rgb]{0.753,0.753,0.753}}0.006 & {\cellcolor[rgb]{0.753,0.753,0.753}}0.609 & {\cellcolor[rgb]{0.753,0.753,0.753}}0.551 \\ \cline{1-2}\cline{4-9}\cline{11-16}\cline{18-23}
\multirow{2}{*}{$\Delta \tau_{peak}$} & Val &  & 1.032 & 2 & 26 & 0.516 & 15.069 & 0.000*** &  & 5.694 & 1 & 13 & 5.694 & 74.026 & 0.000*** &  & 0.667 & 2 & 26 & 0.333 & 11.752 & \multicolumn{1}{l}{0.000***} \\
 & {\cellcolor[rgb]{0.753,0.753,0.753}}Gen &  & {\cellcolor[rgb]{0.753,0.753,0.753}}0.265 & {\cellcolor[rgb]{0.753,0.753,0.753}}2 & {\cellcolor[rgb]{0.753,0.753,0.753}}26 & {\cellcolor[rgb]{0.753,0.753,0.753}}0.133 & {\cellcolor[rgb]{0.753,0.753,0.753}}1.95 & \multicolumn{1}{r}{{\cellcolor[rgb]{0.753,0.753,0.753}}0.163} &  & {\cellcolor[rgb]{0.753,0.753,0.753}}2.01 & {\cellcolor[rgb]{0.753,0.753,0.753}}1 & {\cellcolor[rgb]{0.753,0.753,0.753}}13 & {\cellcolor[rgb]{0.753,0.753,0.753}}2.01 & {\cellcolor[rgb]{0.753,0.753,0.753}}24.514 & {\cellcolor[rgb]{0.753,0.753,0.753}}0.000*** &  & {\cellcolor[rgb]{0.753,0.753,0.753}}0.592 & {\cellcolor[rgb]{0.753,0.753,0.753}}2 & {\cellcolor[rgb]{0.753,0.753,0.753}}26 & {\cellcolor[rgb]{0.753,0.753,0.753}}0.296 & {\cellcolor[rgb]{0.753,0.753,0.753}}8.785 & \multicolumn{1}{l}{{\cellcolor[rgb]{0.753,0.753,0.753}}0.005**} \\ \cline{1-2}\cline{4-9}\cline{11-16}\cline{18-23}
\multirow{2}{*}{$C(\%)$} & Val &  & 311.725 & 2 & 26 & 155.863 & 18.237 & 0.000*** &  & 88.262 & 1 & 13 & 88.262 & 4.683 & \multicolumn{1}{r}{0.05} &  & 15.794 & 2 & 26 & 7.897 & 0.914 & 0.413 \\
 & {\cellcolor[rgb]{0.753,0.753,0.753}}Gen &  & {\cellcolor[rgb]{0.753,0.753,0.753}}1200.068 & {\cellcolor[rgb]{0.753,0.753,0.753}}2 & {\cellcolor[rgb]{0.753,0.753,0.753}}26 & {\cellcolor[rgb]{0.753,0.753,0.753}}600.034 & {\cellcolor[rgb]{0.753,0.753,0.753}}18.429 & {\cellcolor[rgb]{0.753,0.753,0.753}}0.000*** &  & {\cellcolor[rgb]{0.753,0.753,0.753}}295.573 & {\cellcolor[rgb]{0.753,0.753,0.753}}1 & {\cellcolor[rgb]{0.753,0.753,0.753}}13 & {\cellcolor[rgb]{0.753,0.753,0.753}}295.573 & {\cellcolor[rgb]{0.753,0.753,0.753}}17.296 & {\cellcolor[rgb]{0.753,0.753,0.753}}0.001** &  & {\cellcolor[rgb]{0.753,0.753,0.753}}23.504 & {\cellcolor[rgb]{0.753,0.753,0.753}}2 & {\cellcolor[rgb]{0.753,0.753,0.753}}26 & {\cellcolor[rgb]{0.753,0.753,0.753}}11.752 & {\cellcolor[rgb]{0.753,0.753,0.753}}1.508 & {\cellcolor[rgb]{0.753,0.753,0.753}}0.24 \\ \hhline{--~------~------~------}
\end{tabular}
\\[6pt]
\begin{tabular}{cccrcllrcllrcrlrcl} \cline{4-18}
 &  &  & \multicolumn{15}{c}{Post-Hoc} \\ \cline{4-18}
 &  &  & \multicolumn{3}{c}{Persona 1-2} & \multicolumn{1}{c}{} & \multicolumn{3}{c}{Persona 1-3} & \multicolumn{1}{c}{} & \multicolumn{3}{c}{Persona 2-3} & \multicolumn{1}{c}{} & \multicolumn{3}{c}{LUT-TPGMM} \\ \cline{1-2}\cline{4-6}\cline{8-10}\cline{12-14}\cline{16-18}
Measure & Case &  & \multicolumn{1}{c}{T} & df & \multicolumn{1}{c}{P-val} & \multicolumn{1}{c}{} & \multicolumn{1}{c}{T} & df & \multicolumn{1}{c}{P-val} & \multicolumn{1}{c}{} & \multicolumn{1}{c}{T} & df & \multicolumn{1}{c}{P-val} & \multicolumn{1}{c}{} & \multicolumn{1}{c}{T} & df & \multicolumn{1}{c}{P-val} \\ \cline{1-2}\cline{4-6}\cline{8-10}\cline{12-14}\cline{16-18}
\multirow{2}{*}{$\varepsilon$} & Val &  & -4.757 & 13 & 0.001** &  & -5.599 & 13 & 0.000*** &  & 0.605 & 13 & 0.555 &  & -2.761 & 13 & 0.016* \\
 & {\cellcolor[rgb]{0.753,0.753,0.753}}Gen &  & {\cellcolor[rgb]{0.753,0.753,0.753}}3.902 & {\cellcolor[rgb]{0.753,0.753,0.753}}13 & {\cellcolor[rgb]{0.753,0.753,0.753}}0.005** &  & {\cellcolor[rgb]{0.753,0.753,0.753}}3.189 & {\cellcolor[rgb]{0.753,0.753,0.753}}13 & {\cellcolor[rgb]{0.753,0.753,0.753}}0.014* &  & {\cellcolor[rgb]{0.753,0.753,0.753}}-0.373 & {\cellcolor[rgb]{0.753,0.753,0.753}}13 & {\cellcolor[rgb]{0.753,0.753,0.753}}0.715 &  & {\cellcolor[rgb]{0.753,0.753,0.753}}2.852 & {\cellcolor[rgb]{0.753,0.753,0.753}}13 & {\cellcolor[rgb]{0.753,0.753,0.753}}0.014* \\ \cline{1-2}\cline{4-6}\cline{8-10}\cline{12-14}\cline{16-18}
\multirow{2}{*}{$\Delta \tau_{peak}$} & Val &  & -4.386 & 13 & 0.001** &  & -4.708 & 13 & 0.001** &  & -0.489 & 13 & 0.633 &  & -8.604 & 13 & 0.000*** \\
 & {\cellcolor[rgb]{0.753,0.753,0.753}}Gen &  & {\cellcolor[rgb]{0.753,0.753,0.753}}- & {\cellcolor[rgb]{0.753,0.753,0.753}}- & {\cellcolor[rgb]{0.753,0.753,0.753}}- &  & {\cellcolor[rgb]{0.753,0.753,0.753}}- & {\cellcolor[rgb]{0.753,0.753,0.753}}- & {\cellcolor[rgb]{0.753,0.753,0.753}}- &  & {\cellcolor[rgb]{0.753,0.753,0.753}}- & {\cellcolor[rgb]{0.753,0.753,0.753}}- & \multicolumn{1}{l}{{\cellcolor[rgb]{0.753,0.753,0.753}}-} &  & {\cellcolor[rgb]{0.753,0.753,0.753}}-4.951 & {\cellcolor[rgb]{0.753,0.753,0.753}}13 & {\cellcolor[rgb]{0.753,0.753,0.753}}0.000*** \\ \cline{1-2}\cline{4-6}\cline{8-10}\cline{12-14}\cline{16-18}
\multirow{2}{*}{$C(\%)$} & Val &  & 6.084 & 13 & 0.000*** &  & 4.306 & 13 & 0.002** &  & -0.937 & 13 & 0.366 &  & - & - & - \\
 & {\cellcolor[rgb]{0.753,0.753,0.753}}Gen &  & {\cellcolor[rgb]{0.753,0.753,0.753}}-6.243 & {\cellcolor[rgb]{0.753,0.753,0.753}}13 & {\cellcolor[rgb]{0.753,0.753,0.753}}0.000*** &  & {\cellcolor[rgb]{0.753,0.753,0.753}}-4.706 & {\cellcolor[rgb]{0.753,0.753,0.753}}13 & {\cellcolor[rgb]{0.753,0.753,0.753}}0.001** &  & {\cellcolor[rgb]{0.753,0.753,0.753}}0.163 & {\cellcolor[rgb]{0.753,0.753,0.753}}13 & {\cellcolor[rgb]{0.753,0.753,0.753}}0.873 &  & {\cellcolor[rgb]{0.753,0.753,0.753}}-4.159 & {\cellcolor[rgb]{0.753,0.753,0.753}}13 & {\cellcolor[rgb]{0.753,0.753,0.753}}0.001** \\ \hhline{--~---~---~---~---}
\end{tabular}
\begin{tablenotes}
\centering
    \footnotesize
    \item $^{*}p<0.05,~^{**}p<0.01,~^{***}p<0.001.$
\end{tablenotes}
\label{stats}
\end{table*}

\begin{figure*}[htbp!]
    \centering
    \includegraphics[width=0.90\linewidth]{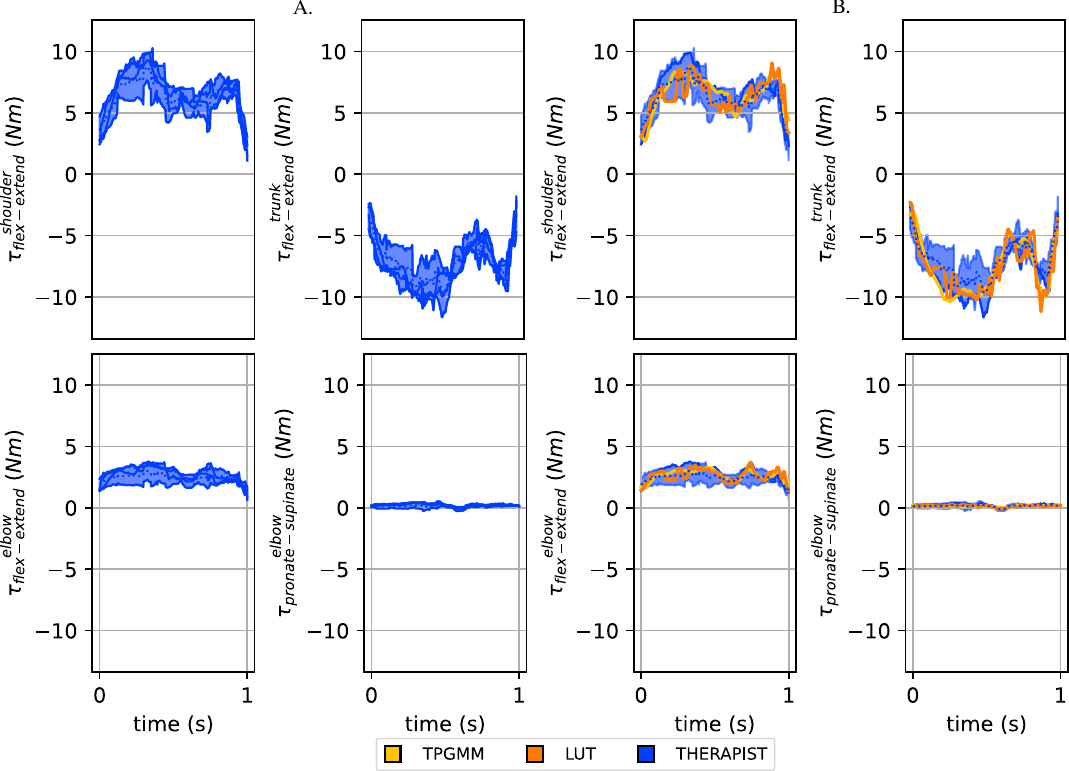}
    \caption{\textcolor{blue}{A:} Variability of therapist behaviour, $\boldsymbol{\tau}_{\zeta = \{18,2,1\}}$ in the trunk, shoulder and elbow. The dotted line represents the mid-range profile $\boldsymbol{m}_{\zeta=\{18,2,1\}}$ and the shaded region represents the behaviour boundaries. \textcolor{blue}{B:} Reconstructed torques from TPGMM (TPGMM) and vector search (LUT) for $\boldsymbol{\tau}_{n = \{18,2,1,3\}}$.} 
    \label{recon}
\end{figure*}
\begin{figure*}[htbp]
    \centering
    \includegraphics[width=1.0\linewidth]{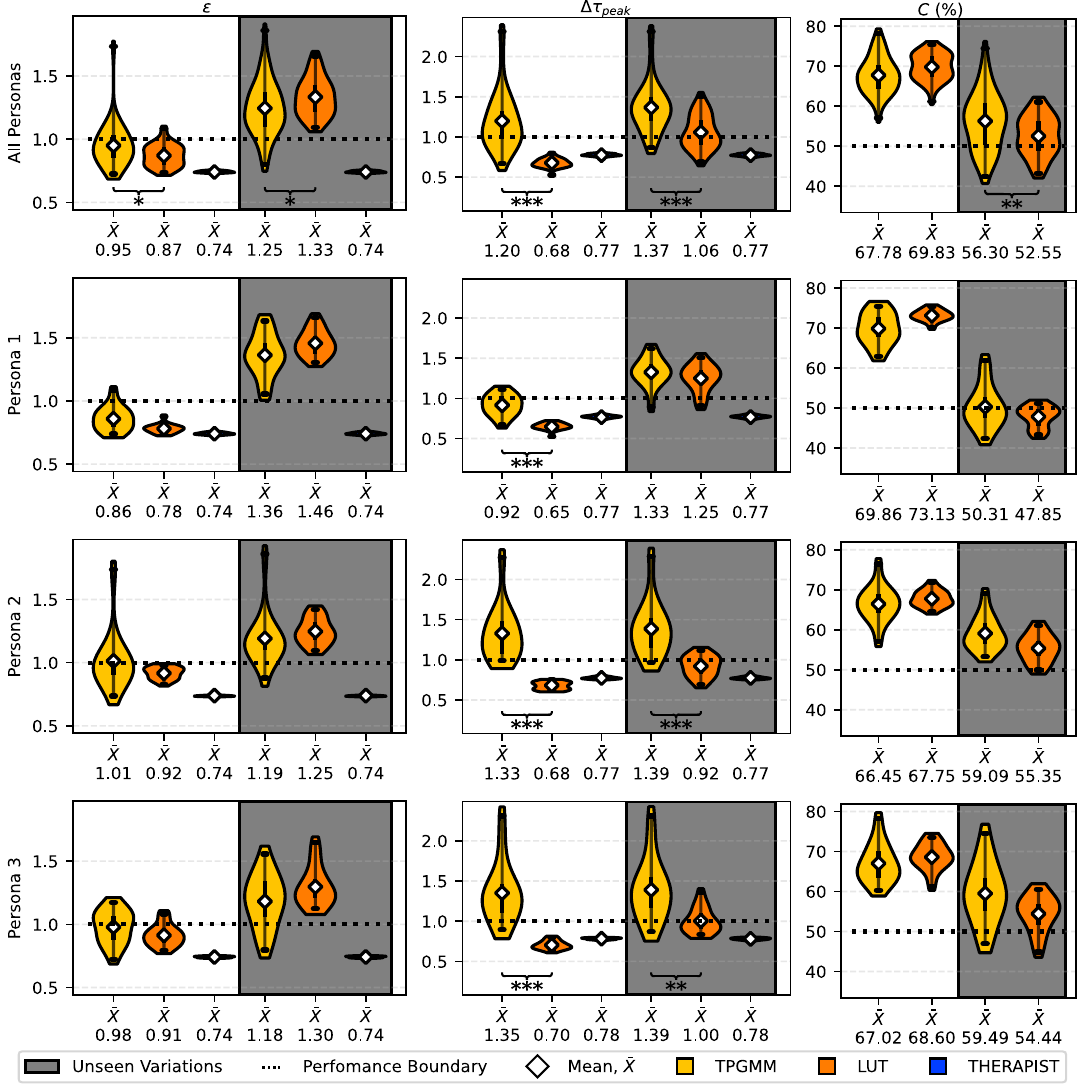}
    \caption{Distribution of outcome measures for both candidate methods between the validation (white background) and generalisation (grey background) cases across patient personas. The therapist's own variability of behaviour is shown as reference.}
    \label{results}
    \vspace{-1.5em}
\end{figure*}

\subsection{Participants and Exclusion}
Twenty four participants without upper-limb injury or known neurological injury ($\#$ of pairs: 12, Age: $25.2\pm0.5$, Height:  $169.4\pm1.8~\mathrm{cm}$, Gender: 20f) were recruited for the study. All participants were instructed to use their right hand when acting out the three patient personas.

For the first five pairs, ``therapists'' were allowed to freely interact with their ``patient''. However, sensor engagement was neglected during the majority of the session, resulting in minimal recorded interactions. Hence, participants after the fifth pair were instructed to restrict their interaction to the 3D-printed cuffs, ensuring that interaction is fully captured by the force sensors. Participants were  given time to adjust their interaction with the ``patient'' before the start of each variation. 

Data from the first five pairs were therefore discarded, leaving data from 14 participants for the full analysis. Data from participant \#20 during the second variation in the third patient persona $\Gamma_{\zeta=\{20,3,2\}}$ encountered a communication error that corrupted two of four trials. Hence, outcome measures obtained from training or test samples from $\Gamma_{\zeta=\{20,3,2\}}$ were excluded from the analysis.

\subsection{Visualisation for $\boldsymbol{\tau}_{\zeta}$ and $\boldsymbol{\hat\tau}_{n}$}
A representative example of the variability of the therapist applied torque for a given variation and the four repetitions ($\boldsymbol{\tau}_{\zeta = \{18,2,1\}}$) is shown in Fig.~\ref{recon} (Left). The corresponding torque reconstructions using TPGMM and LUT approaches ($\boldsymbol{\hat\tau}_{n=\{18,2,1,3\}}$)
are shown in Fig. \ref{recon} (Right). $\tau_{\zeta}$ across patient personas-variations and participants for the shoulder flexion-extension is shown in Fig. S1 in the Supplementary Material. 


\subsection{Distribution of Outcome Measures}
The distribution of the three outcome measures for both candidate methods between the validation and generalisation cases across patient personas are presented on Fig.~\ref{results}.

All outcome measures in both cases are normally distributed ($R^2 > 90$), while $\varepsilon$ and $\Delta\tau_{peak}$ in the validation case appear approximately normal. The deviation from normality of $\varepsilon$ and $\Delta\tau_{peak}$ is traced to a single participant's atypical performance in Persona 2 and 3 for TPGMM. We retained this participant to avoid removing potentially representative samples that reflect genuine population variability. Residuals of each outcome measure for both cases are shown in Fig. S2 in the Supplementary Material.

Outcomes of the repeated-measure ANOVAs and associated post-hoc tests are presented in Table~\ref{stats}. The results of the ANOVA revealed a significant effect of patient personas on all cases-outcome measures except for $\Delta\tau_{\mathrm{peak}}$ in the generalisation case. The choice of candidate methods also significantly affect all cases-outcome measures except for $C$ in the validation case. The interaction effect between Persona and Method was not statistically significant for any outcome measure except for $\Delta\tau_{\mathrm{peak}}$ in both cases, hence the post-hoc pairwise comparisons were not conducted.

\subsubsection{Validation Performance} 
Both methods reconstructed therapist-intended interactions effectively, with both method's $\varepsilon$ distributed below or around $1$ and $C$ distributed around $68\%$. However, TPGMM demonstrated poor fidelity in capturing finer interaction details compared to vector search, evidenced by its $\Delta\tau_{\mathrm{peak}}$ distributed around $1.2$. The choice of method produced a small but significant effect on all outcome measures except for $C$ in the validation case, showing vector search's performance over TPGMM.

\subsubsection{Generalisation Performance} 
Both method's $\varepsilon$ are distributed above $1$ but remained close to the performance boundary, whereas $C$ are distributed around $52-56\%$ across patient personas for both methods. indicating reconstructions trained with minimal training data slightly deviates from the therapist-intended behaviour for unseen variations. TPGMM demonstrated a small but significant advantage over vector search for $\varepsilon$ and $C$, albeit with the same low fidelity in capturing finer interaction details.

\subsubsection{Effects of Patient Persona}
Subsequent post-hocs for patient persona comparisons revealed significant effects when comparing between Persona $1$ to Persona $2$ and $3$, but no significant effects between Persona $2$ and $3$. The results complies with the task complexity design of each patient persona, whereby Persona $1$'s task is simpler, and Persona $2$ and $3$'s tasks are almost equivalent

The increase in task complexity produced opposing effects on method performance between cases. In the validation case, increased complexity from Persona $1$ to Persona $2$ and $3$ significantly degraded performance ($\varepsilon \uparrow$, $C$ $\downarrow$). Conversely, in the generalisation case, increased complexity significantly improved performance ($\varepsilon \downarrow$, $C$ $\uparrow$).

%% file: Sections/5_discussion.tex
\section{Discussion}
In this work, we conducted a preliminary study that extends previous works~\cite{lut_interaction_luciani,lut_interaction_just} on learning physical therapist-patient interaction in upper limb rehabilitation from simple two-point movement to more complex varying task-specific exercises. We learn the relation between patient movement and physical therapist actions using \cite{lut_interaction_luciani}'s vector search algorithm as a benchmark compared against the proposed TPGMM. Both methods are trained on a subset of variations of each task, validated against the same variations with unseen patient movement and generalised against unseen variations. 

Our results demonstrate that both candidate methods reconstructed interactions that stay within the variability of therapist behaviour for seen variations, whereby the vector search algorithm has a slight advantage over TPGMM. In unseen variations, both candidate methods reconstructed interactions that slightly deviated from the variability of therapist behaviour, while under minimal demonstration constraints. TPGMM has a small but significant advantage over the vector search algorithm when reconstructing interaction for unseen variations. However, it should be acknowledged that TPGMM is unable to capture the finer interaction details such as peak torques, of which warrant further investigation.

\subsection{Reduction in Interaction Coverage and Training Samples}
Here, we also validated the use of the vector search algorithm \cite{lut_interaction_luciani} to learn and replicate physical therapist-patient interaction for more complex movement under various conditions, albeit only when reconstructing interaction for variations that is within the training dataset. The difference in performance between TPGMM and the vector search algorithm are also significantly small in the unseen cases.

Several caveats is noted on the performance of the vector search. The selected variations cover $66\%$ of the interaction workspace, whereby samples used for training cover $75\%$ of the seen variations, resulting in 12 training samples out of 24 demonstrations. We suspect the consistency of the therapist's own behaviour across the variations of the task contribute to the performance of the vector search. Hence, further works should explore the effects of reducing variation coverage and training samples, with the aim of reducing the number of demonstrations that the therapist has to provide in the session.

\subsection{Correlation Between Task Complexity and LfD Performance}
A significant relation between task complexity and the performance of LfD when generalising for unseen variations was observed. In \cite{loh_tpgmm_movement_2025}, it was noted that TPGMM's performance over DMPs improves as the task complexity increases, albeit the authors did not analyse the effects of task complexity. Here, we showed that the performance of both candidate methods improves significantly as task complexity increases when generalising for unseen variations. However, this comes at the expense of the performance in the seen variations. While a definite explanation cannot be given due to differences in tasks performed and instructed patient diagnosis, this suggests a trade-off between task complexity and therapist behaviour reproduction in seen and new variations.

\subsection{Limitations and Future works}
In the absence of a system that captures full body interaction between two humans, we constrained interaction to three points of contact (clavicle, upper arm, forearm) to simplify system design. We acknowledge that the constraint significantly alters the interaction performed by the participants compared to conventional therapy sessions. During the pilot study (with ten participants), ``therapists'' were free to interact anywhere, and some interactions were localised to the joints (shoulder, elbow, wrist) which are not sensorised. After enforcing the constraint on interaction points, a similar report was obtained from the remaining participants in that they would preferably support the elbow for reaching motions rather than the upper arm, the wrist for elbow pro-supination, and the fingertips for fine motor control. Six participants report that they would distribute a larger contact surface along a more severe patient's body, even necessitating the presence of two therapists to ensure patient movement stability.

Physiotherapy students were recruited as they were readily available and had not yet developed clinical habits that could bias their engagement with the system developed in Fig. \ref{ubo}. While meaningful and detailed personas were provided, lack of clinical experience could lead to differing interpretation of patient personas between participants, leading to larger interaction variability than in actual practice. 

The LfD frameworks were also evaluated offline on patient kinematics that is unconstrained by any robotic system. In an actual application scenario, the presence of the robot and its physical constraint may affect the interaction of the clinicians in a different way. We also note that the real-time efficacy of the framework with human-in-the loop has yet to be validated, yet TPGMM training lasted between 2 - 4 minutes on a standard CPU depending on task complexity.


These limitations prevents claiming any direct applicability of this findings to actual neuro-rehabilitation settings, nevertheless, the results suggests the suitability of the proposed methods for such scenarios. The next step would be to integrate this framework with an actual rehabilitation robotic system and conduct pilot studies to evaluate its efficacy for rehabilitation post-stroke.

%% file: Biblio/bibliography.bib
@article{tot_efficacy_lee,
  title={Effectiveness of activity-based task-oriented training on upper extremity recovery for adults with stroke: a systematic review},
  author={Lee, Cheng-Yu and Howe, Tsu-Hsin},
  journal={Am. J. Occup. Ther.},
  volume={78},
  number={2},
  pages={7802180070},
  year={2024},
  publisher={The American Occupational Therapy Association, Inc.}
}

@article{tot_efficacy_lang,
  title={Upper-extremity task-specific training after stroke or disability},
  author={Lang, Catherine E and Birkenmeier, Rebecca L},
  journal={Bethesda: AOTA Pr},
  year={2013}
}

@article{dose_efficacy_schneider,
  title={Increasing the amount of usual rehabilitation improves activity after stroke: a systematic review},
  author={Schneider, Emma J and Lannin, Natasha A and Ada, Louise and Schmidt, Julia},
  journal={J. Physiother.},
  volume={62},
  number={4},
  pages={182--187},
  year={2016},
  publisher={Elsevier}
}

@article{dose_efficacy_lang,
  title={Dose and timing in neurorehabilitation: prescribing motor therapy after stroke},
  author={Lang, Catherine E and Lohse, Keith R and Birkenmeier, Rebecca L},
  journal={Curr. Opin. Neurol.},
  volume={28},
  number={6},
  pages={549--555},
  year={2015},
  publisher={LWW}
}

@article{tot_cho,
  title={Effects of robot-assisted task-oriented training on upper limb function and activities of daily living in patients with stroke: A systematic review},
  author={Cho, Yong-Ho and Lee, Myoung-Ho and Cha, Yong-Jun},
  journal={J. Int. Med. Res.},
  volume={54},
  number={5},
  pages={03000605261446490},
  year={2026},
  publisher={SAGE Publications Sage UK: London, England}
}

@article{hhi_johnson,
  title={Therapist-patient interactions in task-oriented stroke therapy can guide robot-patient interactions},
  author={Johnson, Michelle J and Mohan, Mayumi and Mendonca, Rochelle},
  journal={Int. J. Soc. Robot.},
  volume={14},
  number={6},
  pages={1527--1546},
  year={2022},
  publisher={Springer}
}

@inproceedings{vary_tot_chrungoo,
  title={Towards perception driven robot-assisted task-oriented therapy},
  author={Chrungoo, Addwiteey and Shirsat, Priyanka and Johnson, Michelle J},
  booktitle={2015 International Conference on Rehabilitation Robotics (ICORR)},
  pages={660--665},
  year={2015},
}

@article{robot_tot_rozevink,
  title={Effectiveness of task-specific training using assistive devices and task-specific usual care on upper limb performance after stroke: a systematic review and meta-analysis},
  author={Rozevink, Samantha G and Hijmans, Juha M and Horstink, Koen A and van der Sluis, Corry K},
  journal={Disabil. Rehabil.: Assist. Technol.},
  volume={18},
  number={7},
  pages={1245--1258},
  year={2023},
  publisher={Taylor \& Francis}
}

@article{personalise_micera,
  title={Advanced neurotechnologies for the restoration of motor function},
  author={Micera, Silvestro and Caleo, Matteo and Chisari, Carmelo and Hummel, Friedhelm C and Pedrocchi, Alessandra},
  journal={Neuron},
  volume={105},
  number={4},
  pages={604--620},
  year={2020},
  publisher={Elsevier}
}

@article{variability_learning_raviv,
  title={How variability shapes learning and generalization},
  author={Raviv, Limor and Lupyan, Gary and Green, Shawn C},
  journal={Trends Cogn. Sci.},
  volume={26},
  number={6},
  pages={462--483},
  year={2022},
  publisher={Elsevier}
}

@article{variability_learning_A_shea,
  title={Composition of practice: Influence on the retention of motor skills},
  author={Shea, Charles H and Kohl, Robert M},
  journal={Res. Q. Exerc. Sport},
  volume={62},
  number={2},
  pages={187--195},
  year={1991},
  publisher={Taylor \& Francis}
}

@article{variability_learning_B_shea,
  title={Specificity and variability of practice},
  author={Shea, Charles H and Kohl, Robert M},
  journal={Res. Q. Exerc. Sport},
  volume={61},
  number={2},
  pages={169--177},
  year={1990},
  publisher={Taylor \& Francis}
}

@article{tst_pollock,
  title={Interventions for improving upper limb function after stroke},
  author={Pollock, Alex and Farmer, Sybil E and Brady, Marian C and Langhorne, Peter and Mead, Gillian E and Mehrholz, Jan and Van Wijck, Frederike},
  journal={Cochrane Database Syst. Rev.},
  number={11},
  year={2014},
  publisher={John Wiley \& Sons, Ltd}
}

@article{phrhi_vianello,
  title={Robot-Mediated Physical Human--Human Interaction in Rehabilitation: A Position Paper},
  author={Vianello, Lorenzo and Short, Matthew and Manczurowsky, Julia and K{\"u}{\c{c}}{\"u}ktabak, Emek Bar{\i}{\c{s}} and Di Tommaso, Francesco and Noccaro, Alessia and Bandini, Laura and Clark, Shoshana and Fiorenza, Alaina and Lunardini, Francesca and others},
  journal={IEEE Rev. Biomed. Eng.},
  year={2025},
  publisher={IEEE}
}

@article{phrhi_kuccuktabak,
  title={Therapist-exoskeleton-patient interaction for gait therapy},
  author={K{\"u}{\c{c}}{\"u}ktabak, Emek Bar{\i}{\c{s}} and Short, Matthew R and Vianello, Lorenzo and Ludvig, Daniel and Hargrove, Levi and Lynch, Kevin and Pons, Jose},
  journal={Sci. Robot.},
  volume={11},
  number={115},
  pages={eadz9628},
  year={2026},
  publisher={American Association for the Advancement of Science}
}

@article{phrhi_crocher,
  title={Robotic task specific training for upper limb neurorehabilitation: a mixed methods feasibility trial reporting achievable dose},
  author={Crocher, Vincent and Brock, Kim and Simondson, Janine and Klaic, Marlena and Galea, Mary P},
  journal={Disabil. Rehabil.},
  volume={47},
  number={9},
  pages={2349--2357},
  year={2025},
  publisher={Taylor \& Francis}
}

@article{rbt_control_rehab_dalla,
  title={Review on patient-cooperative control strategies for upper-limb rehabilitation exoskeletons},
  author={Dalla Gasperina, Stefano and Roveda, Loris and Pedrocchi, Alessandra and Braghin, Francesco and Gandolla, Marta},
  journal={Front. Robot. AI},
  volume={8},
  pages={745018},
  year={2021},
  publisher={Frontiers Media SA}
}

@article{rbt_control_rehab_baster,
  title={Training modalities in robot-mediated upper limb rehabilitation in stroke: a framework for classification based on a systematic review},
  author={Basteris, Angelo and Nijenhuis, Sharon M and Stienen, Arno HA and Buurke, Jaap H and Prange, Gerdienke B and Amirabdollahian, Farshid},
  journal={J. NeuroEng. Rehabil.},
  volume={11},
  number={1},
  pages={111},
  year={2014},
  publisher={Springer}
}

@phdthesis{lut_interaction_just,
  title={Fusing conventional and robotic rehabilitation therapy},
  author={Just, Fabian},
  year={2020},
  school={ETH Zurich}
}

@article{lut_interaction_luciani,
  title={Therapists’ force-profile teach-and-mimic approach for upper-limb rehabilitation exoskeletons},
  author={Luciani, Beatrice and Sommerhalder, Michael and Gandolla, Marta and Wolf, Peter and Braghin, Francesco and Riener, Robert},
  journal={IEEE Trans. Med. Robot. Bionics},
  volume={6},
  number={4},
  pages={1658--1665},
  year={2024},
  publisher={IEEE}
}

@article{lfd_review_ravichandar,
  title={Recent advances in robot learning from demonstration},
  author={Ravichandar, Harish and Polydoros, Athanasios S and Chernova, Sonia and Billard, Aude},
  journal={Annu. Rev. Control Robot. Auton. Syst.},
  volume={3},
  number={1},
  pages={297--330},
  year={2020},
  publisher={Annual Reviews}
}

@ARTICLE{lauretti_dmp_movement_2017,
  author={Lauretti, Clemente and Cordella, Francesca and Guglielmelli, Eugenio and Zollo, Loredana},
  journal={IEEE Robot. Autom. Lett.}, 
  title={Learning by Demonstration for Planning Activities of Daily Living in Rehabilitation and Assistive Robotics}, 
  year={2017},
  volume={2},
  number={3},
  pages={1375-1382},
  doi={10.1109/LRA.2017.2669369}}

@article{lauretti_dmp_movement_2018,
  title={Learning by demonstration for motion planning of upper-limb exoskeletons},
  author={Lauretti, Clemente and Cordella, Francesca and Ciancio, Anna Lisa and Trigili, Emilio and Catalan, Jose Maria and Badesa, Francisco Javier and Crea, Simona and Pagliara, Silvio Marcello and Sterzi, Silvia and Vitiello, Nicola and others},
  journal={Front. Neurorobotics},
  volume={12},
  pages={5},
  year={2018},
  publisher={Frontiers Media SA}
}

@INPROCEEDINGS{loh_tpgmm_movement_2025,
  author={Loh, Jia Quan and Crocher, Vincent and Oetomo, Denny and Tan, Ying},
  booktitle={2025 International Conference On Rehabilitation Robotics (ICORR)}, 
  title={Can Learning From Demonstration Approaches Encode and Generalise Human Movements for Neurorehabilitation?}, 
  year={2025},
  volume={},
  number={},
  pages={1-7},
  doi={10.1109/ICORR66766.2025.11062990}}

@article{gmm_interaction_maaref,
  title={A gaussian mixture framework for co-operative rehabilitation therapy in assistive impedance-based tasks},
  author={Maaref, Mahdi and Rezazadeh, Alireza and Shamaei, Kimia and Tavakoli, Mahdi},
  journal={IEEE J. Sel. Top. Signal Process.},
  volume={10},
  number={5},
  pages={904--913},
  year={2016},
  publisher={IEEE}
}

@inproceedings{seds_interaction_martinez,
  title={Learning and robotic imitation of therapist's motion and force for post-disability rehabilitation},
  author={Mart{\'\i}nez, Carlos and Tavakoli, Mahdi},
  booktitle={2017 IEEE International Conference on Systems, Man, and Cybernetics (SMC)},
  pages={2225--2230},
  year={2017},
  organization={IEEE}
}

@inproceedings{gmm_interaction_fong,
  title={Kinesthetic teaching of a therapist's behavior to a rehabilitation robot},
  author={Fong, Jason and Tavakoli, Mahdi},
  booktitle={2018 International Symposium on Medical Robotics (ISMR)},
  pages={1--6},
  year={2018},
  organization={IEEE}
}

@INPROCEEDINGS{gmm_interaction_martinez,
  author={Martinez, Carlos Manuel and Fong, Jason and Atashzar, S. Farokh and Tavakoli, Mahdi},
  booktitle={2019 IEEE Global Conference on Signal and Information Processing (GlobalSIP)}, 
  title={Semi-autonomous Robot-assisted Cooperative Therapy Exercises for a Therapist’s Interaction with a Patient}, 
  year={2019},
  volume={},
  number={},
  pages={1-5},
  doi={10.1109/GlobalSIP45357.2019.8969143}}

@article{pff_interaction_najafi,
  title={Using potential field function with a velocity field controller to learn and reproduce the therapist's assistance in robot-assisted rehabilitation},
  author={Najafi, Mohammad and Rossa, Carlos and Adams, Kim and Tavakoli, Mahdi},
  journal={IEEE/ASME Trans. Mechatron.},
  volume={25},
  number={3},
  pages={1622--1633},
  year={2020},
  publisher={IEEE}
}

@article{gmm_interaction_sorkhabadi,
  title={Learning post-stroke gait training strategies by modeling patient-therapist interaction},
  author={Sorkhabadi, Seyed Mostafa Rezayat and Smith, Mason and Khodmbashi, Roozbeh and Lopez, Rachel and Raasch, Melissa and Maruyama, Trent and Kwasnica, Christina and Zhang, Wenlong},
  journal={IEEE Trans. Neural Syst. Rehabil. Eng.},
  volume={31},
  pages={1687--1696},
  year={2023},
  publisher={IEEE}
}

@article{tpgmm_calinon,
  title={A tutorial on task-parameterized movement learning and retrieval},
  author={Calinon, Sylvain},
  journal={Intell. Serv. Robot.},
  volume={9},
  number={1},
  pages={1--29},
  year={2016},
  publisher={Springer}
}

@article{bic_hastie,
  title={The elements of statistical learning: data mining, inference and prediction},
  author={Hastie, Trevor and Tibshirani, Robert and Friedman, Jerome and Franklin, James},
  journal={Math. Intell.},
  volume={27},
  number={2},
  pages={83--85},
  year={2005},
  publisher={Springer}
}

@article{XSENS,
  title={MVN Anatonomical model},
  author={XSENS},
  journal={MVN User Manual},
  year={2025},
  url={https://go.unimelb.edu.au/ff32}
}

@INPROCEEDINGS{dtw_sempena,
  author={Sempena, Samsu and Nur Ulfa Maulidevi and Peb Ruswono Aryan},
  booktitle={Proceedings of the 2011 International Conference on Electrical Engineering and Informatics}, 
  title={Human action recognition using Dynamic Time Warping}, 
  year={2011},
  volume={},
  number={},
  pages={1-5},
  doi={10.1109/ICEEI.2011.6021605}}
